\pdfoutput=1

\documentclass{article}
\usepackage{arxiv}

\usepackage{times}
\usepackage{helvet}
\usepackage{courier}
\usepackage{subfig}
\usepackage{graphicx}
\usepackage{algorithm} 
\usepackage{algpseudocode} 

\usepackage{wrapfig}
\usepackage{changepage}
\usepackage{lscape}
\usepackage{times}
\usepackage{url}
\usepackage{paralist}
\usepackage{booktabs}
\usepackage{lscape}
\usepackage{array}
\usepackage[T1]{fontenc}
\usepackage{hyperref}
\usepackage[all]{hypcap}
\usepackage{amssymb}
\usepackage{soul}
\usepackage{verbatim}
\usepackage{latexsym}
\usepackage{amsfonts}
\usepackage{amsmath}
\usepackage{amssymb}
\usepackage{rotating}
\usepackage{multirow}
\usepackage{multicol}
\usepackage{pstricks,pst-node}
\usepackage{wrapfig}
\usepackage{tabu}
\usepackage{graphics,graphicx}
\usepackage{subfig}
\usepackage{xcolor}
\usepackage{natbib}
\usepackage{algorithmicx}

\usepackage{etoolbox}
\usepackage{tikz}
\usetikzlibrary{tikzmark}
\usetikzlibrary{calc}

\errorcontextlines\maxdimen

\newcommand{\ALGtikzmarkcolor}{black}
\newcommand{\ALGtikzmarkextraindent}{4pt}
\newcommand{\ALGtikzmarkverticaloffsetstart}{-.5ex}
\newcommand{\ALGtikzmarkverticaloffsetend}{-.5ex}
\makeatletter
\newcounter{ALG@tikzmark@tempcnta}

\newcommand\ALG@tikzmark@start{%
    \global\let\ALG@tikzmark@last\ALG@tikzmark@starttext%
    \expandafter\edef\csname ALG@tikzmark@\theALG@nested\endcsname{\theALG@tikzmark@tempcnta}%
    \tikzmark{ALG@tikzmark@start@\csname ALG@tikzmark@\theALG@nested\endcsname}%
    \addtocounter{ALG@tikzmark@tempcnta}{1}%
}

\def\ALG@tikzmark@starttext{start}
\newcommand\ALG@tikzmark@end{%
    \ifx\ALG@tikzmark@last\ALG@tikzmark@starttext
    \else
        \tikzmark{ALG@tikzmark@end@\csname ALG@tikzmark@\theALG@nested\endcsname}%
        \tikz[overlay,remember picture] \draw[\ALGtikzmarkcolor] let \p{S}=($(pic cs:ALG@tikzmark@start@\csname ALG@tikzmark@\theALG@nested\endcsname)+(\ALGtikzmarkextraindent,\ALGtikzmarkverticaloffsetstart)$), \p{E}=($(pic cs:ALG@tikzmark@end@\csname ALG@tikzmark@\theALG@nested\endcsname)+(\ALGtikzmarkextraindent,\ALGtikzmarkverticaloffsetend)$) in (\x{S},\y{S})--(\x{S},\y{E});%
    \fi
    \gdef\ALG@tikzmark@last{end}%
}

\apptocmd{\ALG@beginblock}{\ALG@tikzmark@start}{}{\errmessage{failed to patch}}
\pretocmd{\ALG@endblock}{\ALG@tikzmark@end}{}{\errmessage{failed to patch}}
\makeatother

\newcommand{\argmax}{\ensuremath{\mathrm{argmax}}}
\newcommand{\argmin}{\ensuremath{\mathrm{argmin}}}

\newcommand{\ignore}[1]{}

\newcommand{\RAT}{\ensuremath{\mathbb{R}}}

\renewcommand{\epsilon}{\varepsilon}

\newcommand{\pydgga}{\textsc{PyDGGA}}

\definecolor{verylightgray}{RGB}{150,150,150}
\sethlcolor{verylightgray}

\title{Learning to Optimize Black-Box Functions With Extreme Limits on the Number of Function Evaluations}

\author{ {Carlos Ans\'{o}tegui} \\
	DIEI, Universitat de Lledia, Spain\\
	\texttt{carlos@diei.udl.ca} \\
	\And
	{Meinolf Sellmann} \\
	General Electric, USA\\
	\texttt{meinolf@ge.com}\\
	\And
	{Tapan Shah} \\
	General Electric, USA\\
	\texttt{tapan.shah@ge.com}\\
	\And
	{Kevin Tierney} \\
	Decision and Operation Technologies Group, \\
	Bielefeld University, Germany\\
	\texttt{kevin.tierney@uni-bielefeld.de}\\
}

\renewcommand{\shorttitle}{Learning to Optimize Black-Box Functions With Extreme Limits on the Number \dots}
\hypersetup{
pdftitle={Learning to Optimize Black-Box Functions With Extreme Limits on the Number of Function Evaluations},
pdfsubject={cs.AI},
pdfauthor={Carlos Ansotegui, Meinolf Sellmann, Tapan Shah, Kevin Tierney},
pdfkeywords={black-box optimization, few shot optimization},
}

\begin{document}
\maketitle

\begin{abstract}
We consider black-box optimization in which only an extremely limited number of function evaluations, on the order of around 100, are affordable and the function evaluations must be performed in even fewer batches of a limited number of parallel trials. This is a typical scenario when optimizing variable settings that are very costly to evaluate, for example in the context of simulation-based optimization or machine learning hyperparameterization. We propose an original method that uses established approaches to propose a set of points for each batch and then down-selects from these candidate points to the number of trials that can be run in parallel. The key novelty of our approach lies in the introduction of a hyperparameterized method for down-selecting the number of candidates to the allowed batch-size, which is optimized offline using automated algorithm configuration. We tune this method for black box optimization and then evaluate on classical black box optimization benchmarks. Our results show that it is possible to learn how to combine evaluation points suggested by highly diverse black box optimization methods conditioned on the progress of the optimization. Compared with the state of the art in black box minimization and various other methods specifically geared towards few-shot minimization, we achieve an average reduction of 50\% of normalized cost, which is a highly significant improvement in performance.
\end{abstract}

\section{Introduction}
We consider the situation where we need to optimize the input variables for unconstrained objective functions that are very expensive to evaluate. Furthermore, the objective functions we consider are \textit{black boxes}, meaning we can use no knowledge from the computation of the objective function (such as a derivative) during search. Such objective functions arise when optimizing over simulations, engineering prototypes, and when conducting hyperparameter optimization for machine learning algorithms and optimization solvers. 

Consider the situation where we wish to optimize the geometry of a turbine blade whose effectiveness will be evaluated by running a number of computational fluid dynamics (CFD) simulations over various different scenarios simulating different environmental conditions. Each evaluation of a geometry design requires running the CFD simulations for all scenarios and may take hours or even days to complete.  In such a setting, we may evaluate multiple scenarios and geometries at the same time to reduce the elapsed time until a high quality design is found.  We could, for example, evaluate 5 designs in parallel and run 20 such batches sequentially. In this case, whenever we set up the new trials for the next batch, we carefully choose 5 new geometries based on the designs and results that were obtained in earlier batches. In total, we would evaluate 100 designs, and the total elapsed time to obtain the optimized geometries would take 20 times the time it takes to evaluate one design for one scenario (provided we are able to run all environmental conditions for all 5 candidate designs in parallel).

From an optimization perspective, this setup is extremely challenging. We are to optimize a black box function in which there is only time enough to try a mere 100 input settings, and, on top of that, we have to try 5 inputs in parallel, which implies we cannot learn from the results obtained from the inputs being evaluated in parallel before choosing these inputs. In analogy to the idea of parallel few-shot learning, we refer to this optimization setup as parallel few-shot optimization.

The objective of the work presented in this paper is to develop a hyperparameterized heuristic that can be tuned offline to perform highly effective parallel few-shot optimization. Our approach involves a Monte-Carlo simulation in which we try out different combinations of candidate points from a larger pool of options. The hyperparameters of our approach control a scoring function that guides the creation of batches of candidates, and we select our final batch of candidates based on which candidates appear most often in the sampled batches. It turns out that setting the hyperparameters of our heuristic is itself a type of black-box optimization (BBO) problem that we solve using the algorithm configurator GGA~\citep{GGA++} as in~\cite{hrds} and \cite{hypertabu}. In this way, our heuristic can be customized to a specific domain, although, in this work, we try to make it work generally for BBO functions. A further key advancement of our approach is that our heuristic adjusts the scoring of new candidates to be added to a sampled batch based on the candidates already in the batch, ensuring a balanced exploration and exploitation of the search space.

In the following, we review the literature on BBO and few-shot optimization, introduce our new method, and, finally, provide numerical results on standard BBO benchmarks. 

\section{Related Work}
We split our discussion of the related work into two groups of approaches: candidate generators and candidate selectors. Given a BBO problem, candidate generators propose points that hopefully offer good performance. This group of approaches can be further divided into evolutionary algorithms, grid sampling, and model-based approaches. Candidate selectors choose from a set of candidate points provided by candidate generators, or select which candidate generators should be used to suggest points, following the idea of the no free lunch theorem~\citep{adam2019} that no single approach (in this case a candidate generator) dominates all others. These approaches include multi-armed bandit algorithms and various forms of ensembles.

\subsection{Candidate generators}
Generating candidates has been the focus of the research community for a long time, especially in the field of engineering, where grid sampling approaches and fractional factorial design~\citep{gunst2009} have long been used to suggest designs (points) to be realized in experiments. Fractional factorial design suggests a subset of the cross-product of discrete design choices that has desirably statistical properties. Latin hypercube sampling~\citep{mckay2000} extends this notion to create candidate suggestions from continuous variables. These techniques are widely used when no information is available about the space, i.e., it is not yet possible inference where good points may be located. 

Evolutionary algorithms (EAs) offer simple, non-model-based mechanisms for optimizing black-box functions and have long been used for BBO~\citep{back1993}. EAs initialize a \emph{population} of multiple solutions (candidate points) and \emph{evolve} the population over multiple generations, meaning iterations, by recombining and resampling solutions in population $i$ to form population $i+1$. The search continues until a termination criterion is reached, for example, the total number of evaluations or when the average quality of the population stops improving.

Standard algorithms in the area of EAs include the 1+1 EA~\citep{rechenberg1978evolutionsstrategien,1plus1EA} and differential evolution (DE)~\citep{storn_differential_1997}. These approaches have seen great research advancements over the past years, such as including self-adaptive parameters~\citep{qin_self-adaptive_2005,tanabe_improving_2014} (SADE, L-SHADE), using memory mechanisms~\citep{brest_il-shade_2016} (iL-SHADE), specialized mutation strategies~\citep{brest_single_2017} (jSO), and covariance matrix learning~\citep{awad_ensemble_2017} (L-SHADE-cnEpSin). The covariance matrix can also be used outside of a DE framework. The covariance matrix adaptation evolutionary strategy (CMA-ES)~\citep{CMA-ES} is one of the most successful EAs, and offers a way of sampling solutions from the search space according to a multivariate normal distribution that is iteratively updated according to a maximum likelihood principle.

EAs have also been used for BBO in the area of algorithm configuration (AC). In the AC setting the black-box is a parameterized solver or algorithm whose performance must be optimized over a dataset of representative problem instances, e.g., when solving a delivery problem each instance could represent the set of deliveries each day. The GGA method~\citep{GGA} is a ``gender-based'' genetic algorithm that partitions its population in two. One half is evaluated with a racing mechanism and is the winners are recombined with members from the other half. 

Model-based approaches are techniques that build an internal predictive model based on the performance of the candidates that were chosen in the past. These methods can offer advantages over EA and grid-sampling approaches in their ability to find high quality solutions after a learning phase. However, this comes at the expense of higher computation time. In the area of AC, the GGA++~\citep{GGA++} technique combines EAs with a random forest surrogate that evaluates the quality of multiple candidate recombinations, returning the one that ought to perform the best. 

Bayesian optimization (BO) has become a widespread model-based method for selecting hyperparameters in black-box settings (see, e.g.,~\cite{frazier2018}) and for the AutoML setting~\citep{snoek2012,feurer2019}. BO models a surrogate function, typically by using a Gaussian process model, which estimates the quality of a given parameter selection as well as the uncertainty of that selection. A key ingredient for BO is the choice of an \textit{acquisition function}, which determines how the optimizer selects the next point to explore. 
There are numerous BO variants, thus we only point out the ones most relevant to this work. For few-shot learning, \cite{wistuba2021} propose a deep kernel learning approach to allow for transfer learning. In~\cite{atkinson2020}, a BO approach for few-shot multi-task learning is proposed. Recently, the NeurIPS 2020 conference hosted a challenge for tuning the hyperparameters of machine learning algorithms~\nocite{bbochallenge}[BBO Challenge, 2020], with the HEBO approach~\citep{cowen2020}, emerging as the victor. HEBO is based on BO and selects points from a multi-objective Pareto frontier, as opposed to most BO methods which only consider a single criterion. 

\subsection{Candidate selectors}
The line between candidate generator and candidate selector is not clear cut, indeed even the fractional factorial design method not only suggests candidates (the cross product of all options), but also provides a mechanism for down-selecting. Thus, by candidate selector, we mean methods that can be applied generally, i.e., the candidates input to the method can come from any variety of candidate generators, or the selector could accept/choose candidate generation algorithms, such as in the setting of algorithm selection~\citep{aslib}. In~\cite{liu2020}, which also competed in the challenge described above, an ensemble generation approach for BBO is presented using GPUs. The resulting ensemble uses the Turbo optimizer~\citep{eriksson2019} (itself a candidate selector using a bandit algorithm) and scikit-optimize~\citep{scikitopt}. In~\cite{ye2018}, an ensemble of three approaches is created and a hierarchy is formed to decide which to use to select points.

Multi-armed bandit approaches are a well-known class of candidate selectors. As we consider the case where multiple candidates in each iteration should be selected, combinatorial bandits with semi-bandit feedback (e.g.,~\cite{lattimore18}) are most relevant. These approaches generally assume the order of observations (between batches) is irrelevant, however we note that, in our case, this is not true. For example, some approaches may work better at selecting points in the first few rounds, while others may excel later on or once particular structures are discovered. Contextual bandits~\citep{lu2010} allow for the integration of extra information, such as the current iteration, to be included in arm selection. The CPPL approach of~\cite{mesaoudipaul2020} uses a Placket-Luce model to choose the top-$k$ arms in a contextual setting, but is meant for situations with a richer context vector, such as algorithm selection, rather than BBO candidate selection.

\section{Hyperparameterized Parallel Few-Shot Optimization (HPFSO)}
Having reviewed the dominant methodologies for BBO, we now introduce our new hyperparameterized approach for parallel few-shot optimization {\bf (HPFSO)}. The idea for our approach is simple: When determining the next batch of inputs to be evaluated in parallel, we employ multiple different existing methodologies to first produce a larger set of candidate inputs. The core function we introduce strategically selects inputs from the superset of candidates until the batch-size is reached. The mechanism for performing this reduction is a hyper-configurable heuristic that is learned using an AC algorithm offline. The selected candidates are evaluated in parallel and the results of all of these trials are communicated to all the candidate-generating methods. That is to say, every point generator is also informed about the true function values of candidates it itself did not propose for evaluation. This process is repeated until the total number of iterations is exhausted. In the end, we return the input that resulted in the best overall evaluation.

\subsection{Candidate Generators}
We first require several methods for generating a superset of candidate inputs from which we will select the final batch that will be evaluated in parallel. To generate candidates, we use: 
\begin{itemize}
    \item  Latin Hypercube Sampling ({\bf LHS}): For a requested number of points $k$, partition the domain of each variable into $k$ equal (or, if needed, almost equal) sized intervals. For each variable, permute the $k$ partitions randomly and independently of the other variables. Create the $i$-th point by picking a random value from partition $i$ (in the respective permuted ordering) for each variable.
    \item Bayesian Optimization: Bayesian optimization updates a surrogate model with new point(s) $x_n$ and their evaluated values. Using the surrogate model as response surface,  an acquisition function $\alpha(x)$ is minimized to derive the next query point(s)
    \begin{align*}
    x_{n+1} = \argmin_x {\alpha(x)}
    \end{align*}
    We run the above minimization $k$ times to generate $k$ points, each time with a different seed. 
     \begin{itemize} 
         \item  Gradient Boosting Tree -- Lower Confidence Bound ({\bf GBM-LCB}): We use a GBM as a surrogate model followed by the LCB acquisition  function 
         \begin{align*}
         \alpha(x) = \mu(x)-\kappa \sigma(x),
         \end{align*}
         where $\mu(x)$ is the posterior mean, $\sigma(x)$ is the posterior standard deviation and $\kappa>0$ is the exploration constant. The parameter $\kappa$ adjusts the bias regarding exploration vs. exploitation. In our experiments, we set $\kappa$ to~2.
         \item Modified Random Forest ({\bf GGA++}): We use the surrogate proposed in~\cite{GGA++}, which directly identifies areas of interest instead of forecasting black-box function values. We use local search to generate local minima over this surrogate without using any form of uncertainty estimates to generate points.
    \end{itemize} 
    \item  Covariance Matrix Adaptation: This is an evolutionary approach that samples $k$ points from a Gaussian distribution that is evolved from epoch to epoch. The covariance matrix is adjusted based on the black box evaluations conducted so far~\citep{CMA-ES}. We use the covariance matrix in two ways to generate points:
    \begin{itemize}
        \item For sampling using the current best point as mean of the distribution {\bf(CMA)}.
        \item For sampling using the mean of the best point suggested by the GGA++ surrogate {\bf(CMA-N)}.
    \end{itemize}
     \item  Turbo ({\bf TUR}): This method employs surrogate models for local optimization followed by an implicit multi-armed bandit strategy to allocate the samples among the different local optimization problems~\citep{eriksson2019}.
    \item Recombinations of the best evaluated points and points that were suggested but not selected in earlier epochs: We create a {\em diversity store} of all the points recommended in the previous iterations by BO, covariance matrix adaptation and Turbo, but not evaluated after the down-selection. We create recombinations of these points using two methods:
    \begin{itemize}
    \item \textbf{REP}: We select a random point in the diversity store and use path relinking~\citep{glover1997} to connect it with the best point found so far. 
    We choose the recombination on the path with the minimum value as priced by the GGA++-surrogate.
    \item \textbf{RER}: In a variant of the above method, we use random crossover 1000 times for randomly chosen points from the diversity store. We perform pricing again using the GGA++ surrogate, and suggest the best $k$ points to be evaluated.
    \end{itemize}
   
\end{itemize}
Summarizing, we use 8 candidate point selectors: LHS, GBM-LCB, GGA++, CMA, CMA-N, TUR, REP and RER. We will be using these acronyms going forward. 

\subsection{Sub-selection of Candidates}
\newcommand{\nsimu}{\ensuremath{N}}
\begin{algorithm}[t]
\scriptsize
	\caption{Candidate sub-selection} \label{alg:subsel}
	\begin{algorithmic}[1]
	    \State {\bf Input:} $F$:  set of feature vectors  for $C$ candidates; $w$: feature weights; $B$: \# of final candidates, \nsimu: \# simulations
	    \State {\bf Output:} $S$: Indices of selected candidates
	    \State {\bf Initialize} $S \gets \{\}$, $R \gets \{1,2,\ldots,C\}$
		\While {$|S|<B$} \label{line:mainwhile}
		    \State $Q \gets $ vector of length $C$ of zeros
		    \For {$j$ in $1,2,\ldots, \nsimu$} 
		        \State {$S_b \gets S $}
    		    \While {$|S_b|<B$}
        			\State Update and normalize diversity features in $F$ w.r.t. $S_b$
        			\For {$c \in R\setminus S_b $}
        				\State $f^c \gets F(c)$; \; $s_w(f^c) \gets \frac{1}{1+e^{w^Tf^c}}$ \label{line:wtf}
    				\EndFor
    				\State $s_c \gets s_w(f^c)/  \sum_{c=1}^{C}s_w(f^c) \; \forall c \in R\setminus S_b $
                    \State Sample $k$ from  $R\setminus S_b $  with distribution $\{s_c\}$. \label{line:sample}
                    \State $S_b \gets S_b \cup \{k\}$. \label{line:inc}
                    \State $Q[k] \gets Q[k]+1$
                   \EndWhile
            \EndFor
            \State {$S \gets S \cup \{ \argmax\{ Q \}\}$} \Comment{Choose only one value; break ties uniformly at random.}
		\EndWhile
	\end{algorithmic} 
\end{algorithm}

Having generated a set of candidate points, we next require a method to down-select the number of points to the desired batch size of function evaluations that can be conducted in parallel. This function represents the core of our new methodology and is the primary target for our automated parameter tuning, our goal in this process. 

The selection of candidates is shown in Algorithm~\ref{alg:subsel} and works as follows. In each iteration of the main while loop (line~\ref{line:mainwhile}), we select a candidate for our final selection $S$. We next simulate $\nsimu$ completions of $S$. For each candidate that we could add to the batch, we compute its features apply a logistic regression using a weight vector $w$ that is tuned offline (line~\ref{line:wtf}), providing us with a score as to how good (or bad) candidate $c$ is with respect to the current batch. Note that this is a key part of our contribution; we do not simply take the best $B$ candidates, rather, we ensure that the candidates complement one another according to ``diversity'' features. For example, solutions that are too similar to the solutions in $S_b$ can be penalized through the diversity features to encourage exploration, and will receive a lower score than other candidates, even if they otherwise look promising. Alternatively, the hyperparameter tuner can also decide to favor points that are close to each other to enhance intensification in a region. In any case, given the scores for each candidate, we form a probability distribution from the scores and sample a new candidate for $S_b$ (line~\ref{line:sample}).

\begin{figure}[!tb]
    \centering
    \begin{minipage}{0.45\textwidth}
    \centering
        \includegraphics[width=\linewidth]{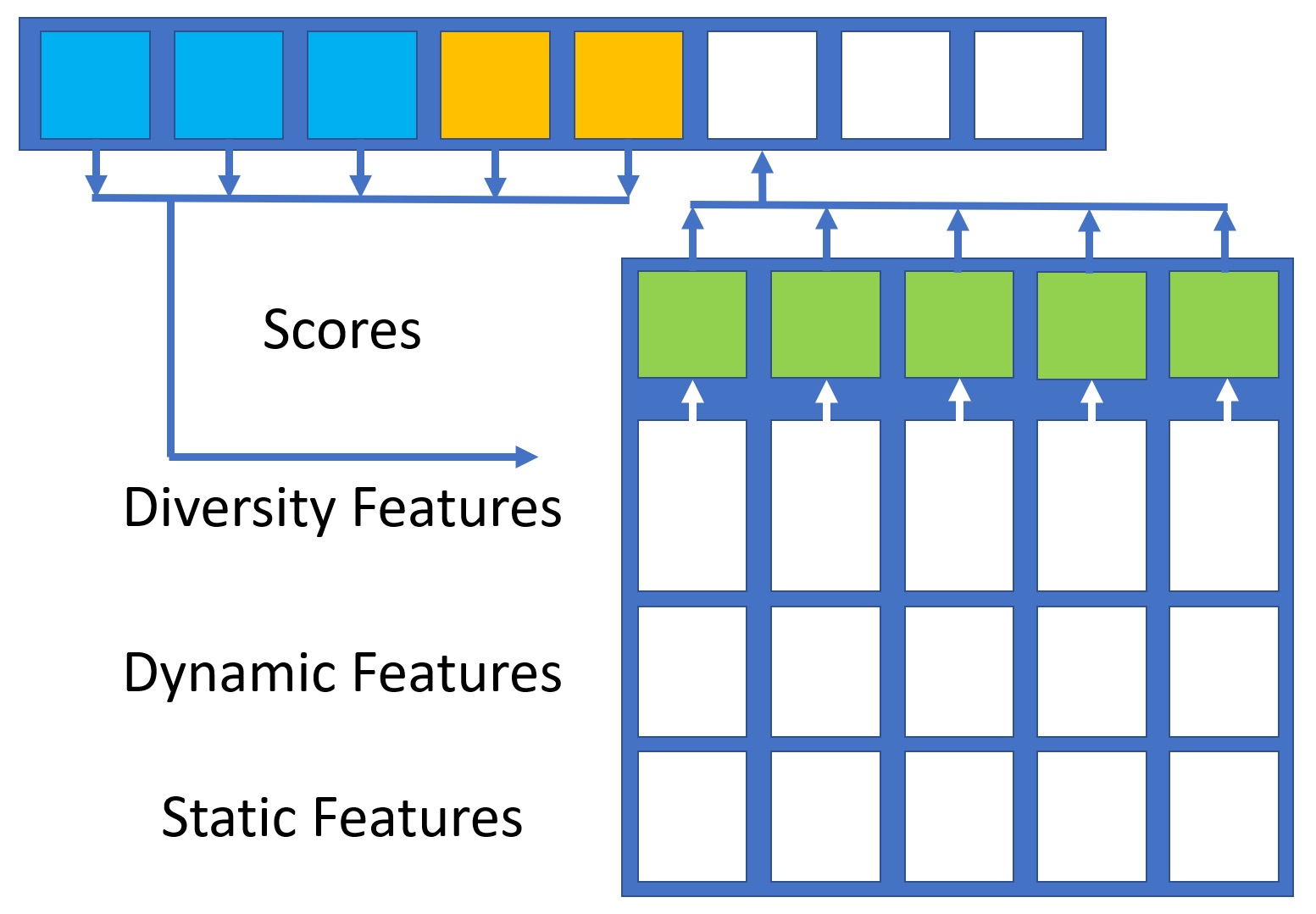}
        \caption{Probability-based selection simulation.}
        \label{fig:sel}
    \end{minipage} \hfill
    \begin{minipage}{0.45\textwidth}
    \centering
        \includegraphics[width=\linewidth]{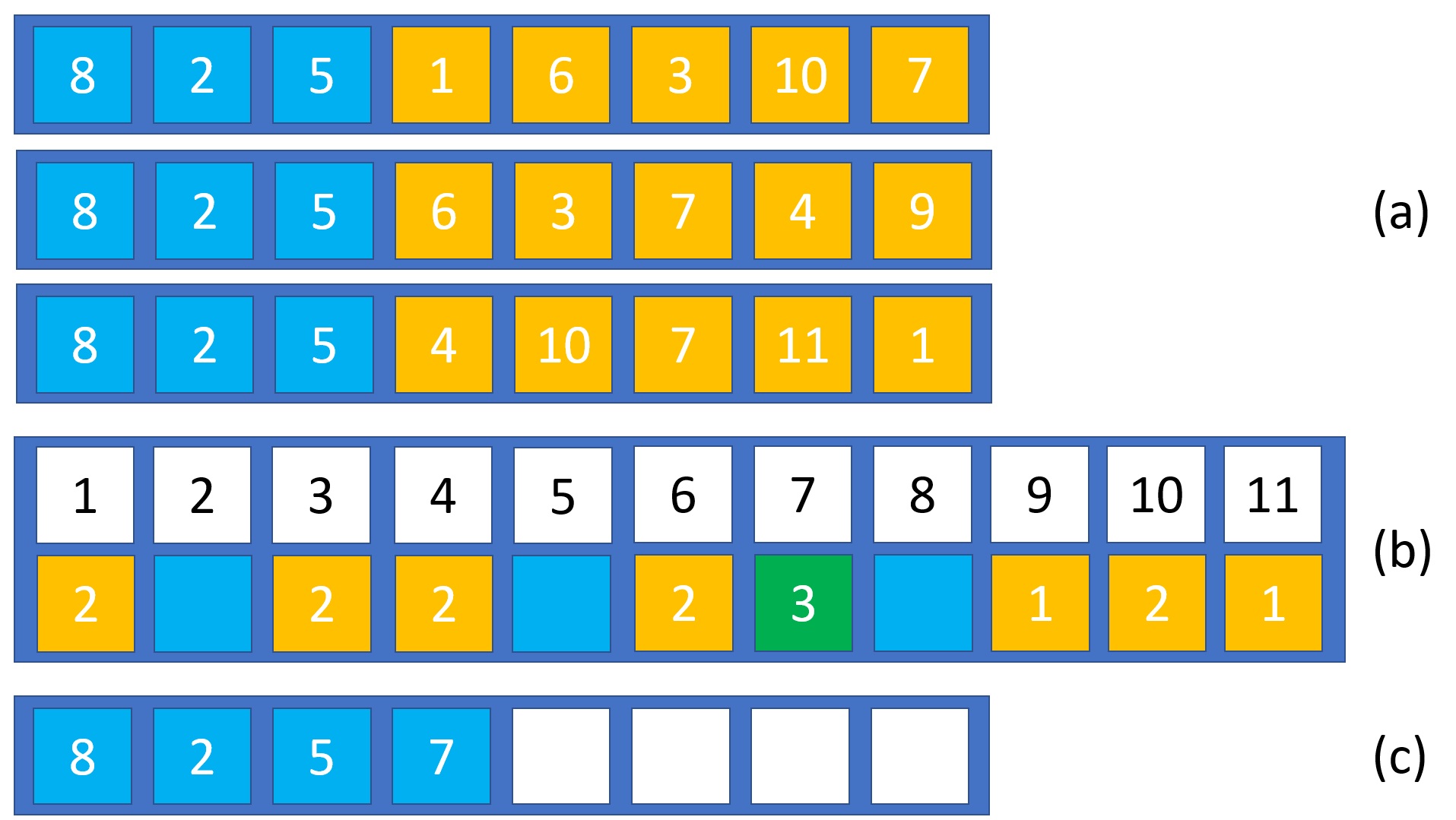}
        \caption{Frequency-based determination of next candidate. (a) Three simulated selections of batches. (b) Frequencies of candidates in simulations. (c) Augmented set of candidates.}
        \label{fig:selfreq}
    \end{minipage}
\end{figure}

Figure~\ref{fig:sel} shows the selection of the next candidate for $S_b$ graphically. The blue squares represent the candidates in $S$, which are fixed in the current simulation. The subsequent two orange cells were selected in the previous two iterations of the current simulation. Now, given three categories of features, which are explained in more detail later, we compute the scores for each of the remaining candidates and choose one at random according to the probability distribution determined by the scores.

Once a candidate has been chosen, we increment a counter for the chosen candidate (line~\ref{line:inc}). Having simulated batches of candidates of the desired size, we then use the frequencies with which the respective candidates appear in the sample batches. The candidate that appears most often gets added to the batch $S$ that we will send to the black box for parallel evaluation, with ties broken uniformly at random. 

Figure~\ref{fig:selfreq} provides a graphical view of the frequency selection of a candidate for $S$. In Figure~\ref{fig:selfreq}(a), we depict again in blue the candidates that are already determined to be part of the final batch of eight. The orange cells show completions of the batch from three simulations. In Figure~\ref{fig:selfreq}(b), we calculate the frequencies (given as $Q$ in the algorithm) with which the different candidates appear in the three sample batches. In green, we show candidate~7, which was selected most often. Finally, in Figure~\ref{fig:selfreq}(c), we provide the augmented partially completed batch with candidate~7. The algorithm will then zero out its frequency table and begin a new round of simulations to fill in the remaining cells with candidate points.

\subsection{Hyperparameterized Scoring Function}

The final piece missing from our approach is the scoring function of candidates that is called during the randomized construction of sample batches. The score for each candidate depends on two pieces of information, the definition of candidate features and the determination of the feature weights $w$. 
We start by listing the features used to characterize each remaining candidate. The features fall into three different categories: 
\begin{enumerate}
    \item \emph{Diversity features}, which rate the candidate in relation to candidates already selected to be part of the sample batch under construction,
    \item \emph{Dynamic features}, which characterize the point with respect to its expected performance, and
    \item \emph{Static features}, which capture the current state of the BBO as a whole.
\end{enumerate}

\subsubsection{Diversity Features:} The first set of features considers how different the candidate is with respect to three different sets of other candidates. These three sets are 1. the set of points that were already evaluated by the black box function in earlier epochs, 2. the set of points already included in the sample batch under construction (denoted with the blue and orange colors above), and 3. the subset of candidates that are already part of the current sample batch and that were generated by the same point generator as the point whose features we are computing. For each of these three sets, we compute the vector of distances of the candidate to each point in the respective set. To turn this vector into a fixed number of features, we then compute some summary statistics over these three vectors: the mean, the minimum, the maximum, and the variance. 

\subsubsection{Dynamic Features:} The second set of features characterizes each candidate in terms of its origin and its expected performance. In particular, we assess the following. What percentage of points evaluated so far were generated by the same point generator as the candidate (if there are none we set the all following values to 0)? Then, based on the vector of function values of these points that were generated by the same generator earlier, what was their average objective value? What was the minimum value? What was the standard deviation? Then, also for these related points evaluated earlier, what was the average deviation of the anticipated objective value from the true objective value?
Next, based on the GBM used as a surrogate by some of the point generators, what is the expected objective value of the candidate? What is the probability that this candidate will improve over the current minimum? And, finally, what is the uncertainty of that probability?

\subsubsection{Static Features:} The last set of features considers the general state of the optimization in relation to the respective candidate. We track the following. What method was used to generate the candidate (this is one-hot encoded, so there are as many of these features as point generator methods)? And, finally, what is the ratio of epochs remaining in the optimization?

\smallskip

After computing all features above for all remaining candidates in each step of building a sample batch, before applying the logistic scoring function, we normalize the diversity and the dynamic features such that the range of each feature is 0 to 1 over the set of all candidates. That is to say, after normalization, for each distance and each dynamic feature, there exists a candidate for which the feature is 0 and another candidate for which the feature is 1 (unless when all feature values are identical, in which case all are set to 0), and all feature values are in $[0,1]$.

\subsubsection{Hyperparameters:} Finally, we need to determine the feature weights $w\in\RAT^n$ to fully define the scoring function. We use an algorithm configurator to determine the hyperparameters, as was previously proposed in~\citep{hrds,hypertabu}, in which AC is used for determining weights of linear regressions in a reactive search.
We tune on a set of 43 black box optimization problems from the 2160 problems in~\cite{hansen2020cocoplat}. As is the practice in machine learning, the functions we tune the method for are different from the functions we evaluate the performance of the resulting method on in the following section. Our test set consists of 157 additional problems sampled at random from the same benchmark.

\medskip

\section{Numerical Results}
We study the performance and scaling behavior of the approach developed in the prior sections by applying it to the established standard benchmarks from the black-box optimization community. 

\subsection{Experimental Setup}
\subsubsection{Benchmark} 
We use the Python API provided by the Comparing Continuous Optimizers (COCO) platform~\citep{hansen2020cocoplat} to generate BBO training and evaluation instances. The framework provides both single objective and multi-objective BBO functions. We use the noiseless BBO functions in our experiments. The suite has 2160 optimization problems (using the standard dimensions from COCO), each specified by a fixed value of identifier, dimension and instance number. We randomly select 2\% (43) problems for  training the hyperparameters and 7.5\% (157) problems for testing.

As each test instance works on its own scale, we normalize the solution values obtained by applying a linear transformation such that the best algorithm's solution value is zero and the worst is one. The algorithms considered for this normalization are all individual point generators as well as the state-of-the-art black box optimizers, but not sub-optimal HFPSO parameterizations, or solutions obtained when conducting different numbers of epochs. Note that, for the latter, values lower than zero or greater than one are therefore possible.

\subsubsection{Configuration of HPFSO's Hyperparameters} 
We tune HPFSO using \pydgga{}~\citep{DGGA}, which is an enhanced version of the GGA++~\citep{GGA++} algorithm configurator~\citep{boosting_amai21} written in the Python language. We tune for 50 wall-clock hours on 80 cores. 

\subsubsection{Contenders} To assess how the novel approach compares to the state of the art we compare with the following approaches: CMA-ES, HEBO~\citep{cowen2020}, and multiple differential evolutionary methods, in particular DE~\citep{storn_differential_1997}, SADE~\citep{qin_self-adaptive_2005}, SHADE~\citep{tanabe_success-history_2013}, L-SHADE~\citep{tanabe_improving_2014} iL-SHADE~\citep{brest_il-shade_2016}, jSO~\citep{brest_single_2017} and L-SHADE-cnEpSin~\citep{awad_ensemble_2017}. We use open open source Python implementations for CMA~\citep{hansen_cma-espycma_2019}, HEBO~\citep{cowen2020} and the DE methods~\citep{ramon_xkuzzpyade_2021}.

\subsubsection{Compute Environment}  All the algorithms were run on a cluster of 80 Intel (R) Xeon CPU E5-2698, 2.20 GHz servers with 2 threads per core, an x86\_64 architecture and the Ubuntu 4.4.0-142 operating system. All the solvers are executed in Python 3.7.

\subsection{Effectiveness of Hyperparameter Tuning}
We begin our study by conducting experiments designed to assess the effectiveness of the hyperparameter tuning. In Table~\ref{tab:hyper}, we show the normalized (see Benchmarks) quality of solutions on the test set. 

\begin{table}[!tb]
\centering
\scriptsize
\begin{tabular}{|l|rr|rrrr|}
\hline
\multirow{2}{*}{} & \multicolumn{2}{|c|}{Random} & \multicolumn{4}{c|}{At Generation} \\
\cline{2-7}
&  \multicolumn{1}{c}{A} &  \multicolumn{1}{c|}{B} &  \multicolumn{1}{c}{5} &  \multicolumn{1}{c}{10} &  \multicolumn{1}{c}{20} &  \multicolumn{1}{c|}{40} \\ \hline
\hline
Mean      &     0.198 &     0.191 &   0.108 &    0.088 &    0.085 &    \bf{0.067} \\ \hline
Std       &     0.287 &     0.199 &   0.188 &    0.176 &    0.146 &    \bf{0.115} \\ \hline
Mean/Gen 40 &     2.955 &     2.851 &   1.612 &    1.313 &    1.269 &    \bf{1.000} \\ \hline
Std/Gen 40  &     2.496 &     1.730 &   1.635 &    1.530 &    1.270 &    \bf{1.000} \\ \hline
\end{tabular}

\vspace{10pt}
\caption{Comparison of randomly chosen hyperparameters with the tuned hyperparameters after different generation of tuning by \pydgga{}}
\label{tab:hyper}
\end{table}
We provide the aggregate performance as measured by the arithmetic mean over the normalized values over all test instances. We compare versions of our novel approach that only differ in the hyperparameters used. The first two versions apply two different random parameterizations (Random A and B). \pydgga{} is based on a genetic algorithm, thus, after each generation, it provides the best performing parameters for HPFSO in that generation. We provide the test performance of the parameterizations found in generation 5, 10, 20 and 40, respectively. Note that the parameters at generation 40 are the last parameters output by \pydgga{}.

We observe that tuning is indeed effective for this method. A priori it was not certain that stochastically tying the static, dynamic, and diversity features to the decision as to which points are selected would have a significant effect on performance at all. Moreover, nor was it certain that we would be able to learn effectively how to skew these stochastic decisions so as to improve algorithm performance. As we can see, however, both of these were possible and this leads to an improvement of about a factor of three in normalized performance over random parameters. At the same time, we also observe a significant drop in variability. As the method gets better at optimizing the functions, the standard deviation also drops, which tells us that the tuning did not lead to outstanding performance on just some instances at the cost of doing a very poor job on others.

\subsection{Importance of the Selection Procedure}
Next, we quantify the impact of the main contribution of our approach, namely the sub-selection procedure. To this end, we compare with each of the 8 point generators included in HPFSO, in isolation, as well as the performance of an approach that employs all point generators and then randomly selects points from the pool that was generated (RAND), and another approach that also uses all 8 point generators and compiles a set of 8 candidate points to be evaluated by selecting the best point (as evaluated by the GBM surrogate) from each point generator (best per method - BPM).

\begin{table}[b]
\centering
\scriptsize
\begin{tabular}{|l|rrrrrrrrrrr|}
\hline
 &    LHS &  CMA-N &    RER &    REP &    TUR &  GGA++ &  GBM-LCB &  CMA &   RAND &    BPM &  HPFSO \\ \hline
\hline
Mean       &  0.606 &  0.612 &  0.549 &  0.543 &  0.534 &  0.521 &    0.341 &  0.142 &  0.266 &  0.269 &  {\bf 0.067} \\ \hline
Std        &  0.304 &  0.357 &  0.336 &  0.311 &  0.304 &  0.307 &    0.317 &  0.207 &  0.228 &  0.252 &  {\bf 0.115} \\ \hline
Mean/HPFSO &  9.182 &  9.273 &  8.318 &  8.227 &  8.091 &  7.894 &    5.167 &  2.152 &  4.030 &  4.076 &  {\bf 1.000} \\ \hline
Std/HPFSO  &  2.667 &  3.132 &  2.947 &  2.728 &  2.667 &  2.693 &    2.781 &  1.816 &  2.000 &  2.211 &  {\bf 1.000} \\ \hline
\end{tabular}

\vspace{10pt}
\caption{Comparison of HPFSO with individual point generators, the best point per method (BPM), and randomly sub-selected points (RAND)}\label{tab:selimp}
\end{table}

In Table~\ref{tab:selimp}, we show the aggregate performances, again as measured by the means of the normalized solution qualities. We observe that the best individual point generator is CMA-ES, followed by BO based on a GBM as surrogate using LCB as acquisition function. Given that CMA-ES is the best point generator in isolation, it may not surprise that the method also beats a random sub-selection of points over all points generated (RAND). What may be less expected is that the best point generator (CMA) in isolation also performs about two times better than choosing the batch consisting of the best points provided by each point generator in each epoch (BPM). And, in fact, this shows the difficulty of the challenge our new method must overcome, as combining the respective strengths of the different point generators appears all but straight forward.

However, when comparing HPFSO with BPM, we nevertheless see that the new method introduced in this paper does manage to orchestrate the individual point generators effectively. On average, HPFSO leads to solutions that incur less than half the normalized cost than any of the point-generation methods employed internally. Moreover, we also observe that the down-selection method works with much greater robustness. The standard deviation is a factor 1.8 lower than that of any competing method.

\begin{wrapfigure}{l}{0.5\textwidth}
\centering
\includegraphics[width=\linewidth]{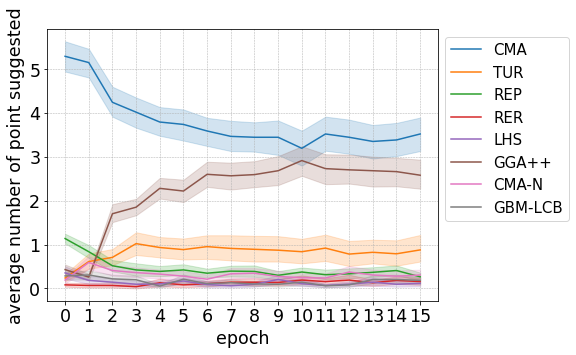}
\caption{Avg. number of points per epoch for each candidate method over test instances
}
\label{fig:genEpoch}
\end{wrapfigure}
In Figure~\ref{fig:genEpoch}, we show the average number of points selected from each point generator as a function over epochs.   We observe that points generated by CMA-ES (CMA) are generally favored by our method, which fills between 40\% and 60\% of the batches with points generated by that method, where the ratio starts at around 60\% and drops gradually down to 40\% in later epochs. Curiously, we also see that CMA-ES is favored in the very first epoch when clearly no learning of the covariant influence of the input variables could have taken place yet. Note that our method would have had the option to employ latin hypercube sampling  instead, but did not. What this tells us is that it appears to be better, even in the very beginning, to gain deeper information in one region of the search space rather than distributing the initial points. We suspect that the superiority of this strategy is the result of the very limited number of black box function evaluations that can be afforded.

Most of the remaining points selected are generated by BO using the GGA++ surrogate, which ramps up quickly to 20\% after two epochs and then steadily grows to about 35\%. This behavior makes intuitive sense, as the GGA++ surrogate is designed to identify favored regions quickly, but does require at least some training data to become effective. The points generated by the Turbo method, which also employs BO internally, show a similar dynamic. After two epochs, about one in eight points is selected from this method, and this ratio stays very steady from then on. 

\begin{table}[b]
\centering
\scriptsize
\begin{tabular}{|l|rrrrrrrrrr|}
\hline
 &  HPFSO &   HEBO &    CMA &  iL-SHADE &   SADE &  SHADE &  L-SHADE &    jSO &  LSHADEcnEpSin &     DE \\ \hline
\hline
 Mean       &  {\bf 0.067} &  0.135 &  0.142 &    0.303 &  0.343 &  0.365 &   0.379 &  0.384 &          0.410 &  0.410 \\ \hline
Std        &  {\bf 0.115} &  0.199 &  0.207 &    0.259 &  0.274 &  0.302 &   0.325 &  0.321 &          0.313 &  0.326 \\ \hline
Mean/HPFSO &  {\bf 1.000} &  2.045 &  2.152 &    4.591 &  5.197 &  5.530 &   5.742 &  5.818 &          6.212 &  6.212 \\ \hline
Std/HPFSO  &  {\bf 1.000} &  1.746 &  1.816 &    2.272 &  2.404 &  2.649 &   2.851 &  2.816 &          2.746 &  2.860 \\ \hline
\end{tabular}

\vspace{10pt}
\caption{Comparison of HPFSO with state-of-art methods}
\label{tab:stateart}
\end{table}

Less intuitive is the strategy to choose a point generated by recombination via path relinking (REP) in the first epoch. We assume that the tuner ``learned'' to add a random point to the batch so as to gauge whether CMA-ES is not searching in a completely hopeless region. However, after two epochs, the influence of REP dwindles down to the same level of influence that all other point generators are afforded by the method. In all epochs, at most 0.3 points from all other point generators are selected on average.

\subsection{Comparison With the State of the Art}

\begin{wrapfigure}{R}{0.5\textwidth}
\centering
\includegraphics[width=\linewidth]{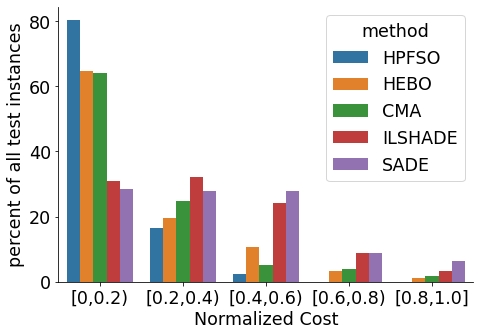}
\caption{Distribution of normalized function values for HPFSO and nearest competitors
}
\end{wrapfigure}

Finally, we provide a comparison with some of the best performing black-box optimization algorithms to date as well as the recent winner from the 2020 NeurIPS Black Box Challenge, HEBO. Note that the established black-box optimizers (such as CMA-ES) have been tuned for the very benchmark we consider, but were not specifically designed to work well for such an extreme limit on the number of function evaluations. HEBO, on the other hand, was developed for the BBO challenge where the objective was to optimize hyperparameters of machine learning algorithms by testing sets of hyperparameters in 16 epochs of 8 samples per epochs. 

In Table~\ref{tab:stateart}, we show the aggregate normalized performance of comparing HPFSO with other competitors. In Figure~\ref{fig:distribution}, we also show the histogram of the normalized function values evaluated by HPFSO and its nearest competitors. 

We observe that CMA-ES and HEBO are the closest contenders, but nonetheless produce solutions that have normalized costs over two times the quality produced by HPFSO. We use a Wilcoxon signed rank test against our data and find that the $p$-value for the hypothesis that HEBO outperforms HPFSO is less than 0.55\%, which allows us to refute this hypothesis with statistical significance.

In Figure~\ref{fig:distribution}, we see that on 80\% of our test instances HPFSO yields a solution that is close to the best performing method for that instance: the normalized function value is between 0 and 0.2. CMA-ES and HEBO also perform relatively well, but still notably worse than HPFSO. We also see that on none of the test instances, HPFSO fails completely. Only on less than 3\% of the test instances, the normalized function value exceeds 0.4, and it is never over 0.6. This again shows the robustness of HPFSO. As we see in Table~\ref{tab:stateart}, HEBO is the closest contender in terms of variability, but even it exhibits a standard deviation that is over 1.7 times larger than that of HPFSO.

\label{fig:distribution}
We also investigate how different setups regarding the number of epochs would affect these results. In Table~\ref{tab:epochs_batch}, we vary the number of epochs between 4 and 24. We use the same batch size of 8 points per parallel trial as before, with the exception of the DE methods which do not allow us to specify parallel trials, so we give these methods a competitive advantage by allowing them to conduct 8 times the number of epochs with one trial per epoch.

Note that HPFSO is trained exclusively using runs with 16 epochs. Nevertheless, the method produces the best results across the board, from optimizing the black box when only 32 function evaluations are allowed in 4 epochs of 8, to an optimization where 192 function evaluations can be afforded in 24 epochs with 8 parallel trials each. This shows that, while the training is targeting a specific number of function evaluations overall, the parameters learned do generalize to a range of other setting as well.

\begin{table}[t]
\scriptsize
\centering
\begin{tabular}{|c|cc|cc|cc|cc|cc|}
\hline
        Epochs & \multicolumn{2}{c|}{4} & \multicolumn{2}{c|}{8} & \multicolumn{2}{c|}{12} & \multicolumn{2}{c|}{16} & \multicolumn{2}{c|}{24}  \\ \hline
               & Mean & Std & Mean & Std & Mean & Std & Mean &Std & Mean & Std  \\ \hline
\hline
         HPFSO &         {\bf 0.419} &         {\bf0.350} &          {\bf0.217} &         0.245 &          {\bf0.137} &         0.201 &          {\bf0.067} &         {\bf0.115} &         {\bf-0.068} &        1.123       \\ \hline
           CMA &          0.477 &         0.408 &          0.265 &         {\bf0.219} &          0.177 &         {\bf0.189} &          0.142 &         0.207 &         -0.031 &         1.080       \\ \hline
          HEBO &          0.825 &         2.434 &          0.345 &         0.570 &          0.200 &         0.252 &          0.135 &         0.199 &          0.073 &         {\bf 0.183}      \\ \hline
       ILSHADE &          1.844 &        11.54 &          0.512 &         0.328 &          0.349 &         0.262 &          0.303 &         0.259 &          0.230 &         0.212  \\ \hline
       GBM-LCB &          0.670 &         0.481 &          0.450 &         0.351 &          0.365 &         0.320 &          0.341 &         0.317 &          0.251 &         0.255 \\ \hline
        LSHADE &         30.32 &       366.5 &          1.555 &        12.19 &          0.532 &         1.120 &          0.379 &         0.325 &          0.299 &         0.278  \\ \hline
          SADE &          0.811 &         1.385 &          0.672 &         2.061 &          0.478 &         0.478 &          0.343 &         0.274 &          0.303 &         0.288  \\ \hline
           jSO &         19.08 &       222.9&          9.327 &       108.6 &          9.683 &       115.9 &          0.384 &         0.321 &          0.322 &         0.323  \\ \hline
 LSAHDECNEP* &         39.80 &       487.3 &          1.530 &        12.197 &          0.521 &         0.893 &          0.410 &         0.313 &          0.325 &         0.280  \\ \hline
         SHADE &         23.30 &       281.6 &          0.635 &         1.560 &          0.460 &         0.414 &          0.365 &         0.302 &          0.334 &         0.288  \\ \hline
            DE &          2.227 &        16.92 &          5.708 &        61.93 &          1.173 &         8.582 &          0.410 &         0.326 &          0.370 &         0.300  \\ \hline
           REP &          1.246 &         1.709 &          0.818 &         0.807 &          0.691 &         0.748 &          0.543 &         0.311 &          0.483 &         0.311  \\ \hline
           RER &          8.014 &        85.32 &          0.710 &         0.543 &          0.600 &         0.389 &          0.549 &         0.336 &          0.494 &         0.322  \\ \hline
         GGA++ &          1.177 &         1.776 &          0.761 &         0.503 &          0.648 &         0.425 &          0.521 &         0.307 &          0.510 &         0.322  \\ \hline
           TUR &          1.119 &         1.506 &          0.753 &         0.545 &          0.633 &         0.354 &          0.534 &         0.304 &          0.518 &         0.331 \\ \hline
           LHS &          1.272 &         3.378 &          0.783 &         0.598 &          0.661 &         0.417 &          0.606 &         0.304 &          0.520 &         0.312 \\ \hline
\end{tabular}
\vspace{10pt}
\caption{Mean and standard deviation of normalized performances of different methods when different numbers of epochs are affordable. The batch size is held constant at 8, except for DE methods whose implementations do not allow for parallel trials. For these methods we allow 8 times the number of epochs, with one trial per epoch.}
\label{tab:epochs_batch}
\end{table}
\smallskip

\section{Conclusion}

We considered the problem of optimizing a black box function when only a very limited number of function evaluations is permitted, and these have to be conducted in a given number of epochs with a specified number of parallel evaluations in each epoch. For this setting, we introduced the idea of using a portfolio of candidate point generators and employed a hyperparameterized method to effectively down-select the set of suggested points to the desired batch size. Our experiments showed that our method can be configured effectively by the \pydgga{} algorithm configurator, and that the primary strength of the method is derived from the parameterized, dynamically self-adapting down-selection procedure. Furthermore, we saw that the resulting method significantly outperforms established black box optimization approaches, as well as a recently introduced method particularly designed for black box optimization with extreme limits on the number of function evaluations.

\bibliographystyle{unsrtnat}
\bibliography{myrefs} 

\end{document}